\documentclass[10pt,twocolumn,letterpaper]{article}

\usepackage{wacv}              % To produce the CAMERA-READY version
\usepackage{graphicx}
\usepackage{amsmath}
\usepackage{amssymb}
\usepackage{booktabs}

\usepackage[pagebackref,breaklinks,colorlinks]{hyperref}

\usepackage[capitalize]{cleveref}
\crefname{section}{Sec.}{Secs.}
\Crefname{section}{Section}{Sections}
\Crefname{table}{Table}{Tables}
\crefname{table}{Tab.}{Tabs.}

\def\wacvPaperID{1452} % *** Enter the WACV Paper ID here
\def\confName{WACV}
\def\confYear{2025}

\usepackage{times}
\usepackage{soul}
\usepackage{url}
\usepackage[utf8]{inputenc}
\usepackage{graphicx}
\usepackage{amsmath}
\usepackage{amsthm}
\usepackage{booktabs}
\usepackage{algorithm}
\usepackage{algorithmic}
\usepackage{multirow}
\usepackage[switch]{lineno}
\usepackage[table]{xcolor}

\definecolor{green1}{HTML}{1e8e1b}
\definecolor{purple}{HTML}{7977B8}
\definecolor{bluee}{HTML}{305bdf}
\definecolor{yellow}{HTML}{e5ba2d}
\definecolor{blue1}{HTML}{84abce}
\usepackage{lipsum}
\usepackage{subcaption}
\usepackage{listings}

\usepackage{color,soul}  % For highlighting
\definecolor{ao(english)}{rgb}{0.0, 0.5, 0.0}
\definecolor{green(english)}{rgb}{0.2, 0.6, 0.2}
\definecolor{orange}{rgb}{0.9, 0.3, 0.0}

\usepackage{array}
\newcolumntype{P}[1]{>{\centering\arraybackslash}p{#1}}
\newcolumntype{L}[1]{>{\raggedright\arraybackslash}p{#1}}

\usepackage[capitalize]{cleveref}
\crefname{section}{Sec.}{Secs.}
\Crefname{section}{Section}{Sections}
\Crefname{table}{Table}{Tables}
\crefname{table}{Tab.}{Tabs.}

\begin{document}

%%%%%%%%% TITLE - PLEASE UPDATE
\title{On the Efficacy of Text-Based Input Modalities for Action Anticipation}

\author{Apoorva Beedu$^{1}$\\
% For a paper whose authors are all at the same institution,
% omit the following lines up until the closing ``}''.
% Additional authors and addresses can be added with ``\and'',
% just like the second author.
% To save space, use either the email address or home page, not both
\and
Harish Haresamudram$^{1}$\\
\and
Karan Samel$^{1}$\\
\and
Irfan Essa$^{1,2}$ \\
\and
Georgia Institute of Technology$^{1}$, Google Research$^{2}$\\
{\tt\small \{abeedu3, hharesamudram, ksamel, irfan\}@gatech.edu}
}
\maketitle

\begin{abstract}
Anticipating future actions is a highly challenging task due to the diversity and scale of potential future actions; yet, information from different modalities help narrow down plausible action choices. 
Each modality can provide diverse and often complementary context for the model to learn from. 
While previous multi-modal methods leverage information from modalities such as video and audio, we primarily explore how text descriptions of actions and objects can also lead to more accurate action anticipation by providing additional contextual cues, e.g., about the environment and its contents. 
We propose a Multi-modal Contrastive Anticipative Transformer (M-CAT), a video transformer architecture that jointly learns from multi-modal features and text descriptions of actions and objects. 
We train our model in two stages, where the model first learns to align video clips with  descriptions of future actions, and is subsequently fine-tuned to predict future actions. 
Compared to existing methods, M-CAT has the advantage of learning additional context from two types of text inputs: rich descriptions of future actions during pre-training, and, text descriptions for detected objects and actions during modality feature fusion. 
Through extensive experimental evaluation, we demonstrate that our model outperforms previous methods on the EpicKitchens datasets, and show that using simple text descriptions of actions and objects aid in more effective action anticipation. 
In addition, we examine the impact of object and action information obtained via text, and perform extensive ablations. 
We will release code upon acceptance.
% We evaluate the performance on on three datasets: EpicKitchens-100, EpicKitchens-55 and EGTEA GAZE+; and show that text descriptions do indeed aid in more effective action anticipation. 
\end{abstract}

%-------------------------------------------------------------------------
\section{Introduction}
\begin{figure}[t]
    \centering
    \includegraphics[width=1\columnwidth]{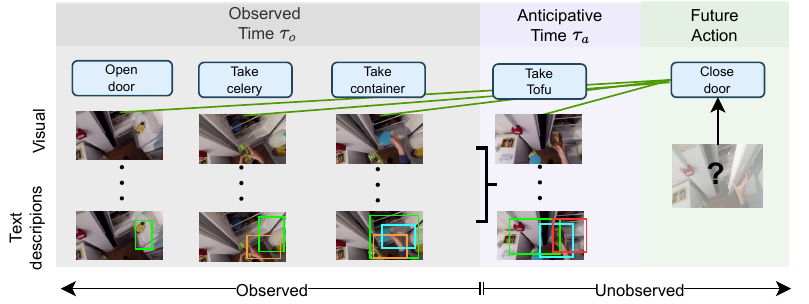}
    \caption{\textbf{Anticipating actions} $\tau_{a}$ seconds after observing information for $\tau_{o}$ seconds using multiple modalities.}
    \label{fig:template}
    \vspace{-1em}
\end{figure}
\label{sec:intro}
Suppose you go to a cafe and order coffee and you see your barista steaming milk, can you predict what they might do next? 
Action anticipation is the task of predicting future actions, using visual cues and data from other modalities such as audio, sensor data, etc.,\ from current and prior observations. 
Predicting future actions is important for many Artificial Intelligence (AI) applications such as autonomous driving~\cite{jain2015car,rasouli2020pedestrian}, assistive robotics~\cite{petkovic2019human,koppula2015anticipating,korbar2018co}, augmented reality, etc.
Although seemingly straightforward for humans, this task is difficult for AI models due to the challenging nature of predicting the future and the wide range of possible actions that the models have to learn. 
Models not only have to detect the action happening at the observed time, but also fuse information from (all available) modalities to anticipate future actions.

Anticipation using only videos (i.e., a single modality) remains challenging, and the availability of additional and complementary modalities is typically advantageous~\cite{girdhar2021anticipative,zhong2023anticipative}.
For instance, an assistive robot can be prepared to help an elderly person if the robot can detect the events leading up to a fall, and anticipate it. 
In addition to video (camera data), audio (sound of the fall, or the person's cry for help), a third person's audio command (``Help the person") etc., can be beneficial.
Accordingly, recent works ~\cite{zhong2023anticipative,girdhar2021anticipative,thakur2023enhancing} have shown that action anticipation greatly benefits from multi-modal training, e.g., using visual and audio cues such as active object detection, and hand-object contact information, ASR etc.
While models are typically trained using modality specific encoders, we examine \emph{if natural language descriptions of actions and objects can be useful for action anticipation}, when employed in addition to other modalities. 
Such descriptions can be highly useful as they can incorporate additional context about the environment and objects required for performing actions, e.g., kitchen vs living room, the utensils utilized, etc., leading to improved action anticipation.
To this end, we leverage the in-context learning capabilities of Large Language Models (LLMs) to generate rich and detailed descriptions of actions and objects.

In this paper, we present a `Multi-modal Contrastive Anticipative Transformer (M-CAT)', that employs a two-stage training process: 
\textit{(i) contrastive pre-training:} where embeddings from videos and other modalities such as optical flow, audio, and natural language descriptions of objects and actions are fused, and contrasted against rich text descriptions of \textit{future actions};
and
\textit{(ii) Action Anticipation:} where the learned embeddings from the modalities are once again fused, and a classifier is trained to predict future actions. 
For both stages, we utilize frozen pre-trained language models (e.g., the CLIP text encoder) to obtain embeddings for text descriptions of object and actions, in lieu of relying on traditional feature extraction methods. 

We study which modalities are more beneficial for action anticipation, and inspect how the accuracy of action recognition for the observed frames affects anticipation. 
As contrastive pre-training typically requires large batch sizes, we also explore alternate avenues of adding more samples during training, specifically for resource constrained setups. 
Finally, we also investigate whether the utilization of an additional self-supervised learning objective can be useful for anticipating actions.   

Therefore, the contributions of our work are: 
\begin{itemize}
    \item We propose a novel approach for predictive video modeling by contrasting multi-modal features against rich text descriptions for future actions, generated using LLMs.
    % \item We investigate whether natural language descriptions of actions and objects can result in improved action anticipation.
    \item We investigate whether using natural language descriptions of actions and objects as an additional modality can result in improved action anticipation.
    \item We improve contrastive pre-training for small batch size capabilities and also introduce an additional self-supervised learning objective.
\end{itemize}

\section{Related Work}
\label{sec:related}
\textbf{Action Anticipation} is the task of predicting future actions after certain time units in a given video clip. 
This task has been explored extensively for third-person videos \cite{abu2018will,gao2017red,huang2014action,jain2016recurrent}.
The release of large-scale egocentric datasets and challenges such as Epic-Kitchen \cite{damen2018scaling,damen2020rescaling} and Ego-4D \cite{grauman2022ego4d} have fast tracked the development for first-person scenarios as well.
To model the temporal progression of past actions, \cite{furnari2020rolling} used a rolling-unrolling-based LSTM network to anticipate actions, such that rolling LSTMs account for the observed video frames, while unrolling LSTMs accounted for the anticipation. 
 \cite{sener2020temporal,sener2021technical} made use of long-range past information by building a multi-scale temporal aggregating framework.
 \cite{thakur2024anticipating,thakur2023enhancing} localize the next active object's position to anticipate actions.
In addition to gathering strong visual features, recent methods have used other visual cues like modeling the environment \cite{nagarajan2020ego} or hand-object contact and activity modeling \cite{dessalene2021forecasting}.
More recently, the use of vision transformers \cite{dosovitskiy2021image} has also been explored. 
While, AVT \cite{girdhar2021anticipative} proposes causal modeling of video frames, and using self-supervision to learn the future frame features, MeMViT \cite{wu2022memsit} perform multi-scale representation of frame features by hierarchically attending the previously cached ``memories".
AFFT \cite{zhong2023anticipative} proposes a fusion method to effectively fuse features from multiple modalities and extend AVT for action anticipation.
 \cite{roy2023predicting}, AntGPT \cite{zhao2023antgpt} and leverages the goal information to reduce the uncertainty in future predictions.
AntGPT \cite{zhao2023antgpt} trains Large Language Models (LLM) to infer goals and model temporal dynamics. 
In contrast, we use pretrained LLMs to generate additional contextual cues about the actions, and create additional text based modalities from objects and actions.

\begin{figure*}[!t]
    \centering
    \includegraphics[width=0.9\linewidth]{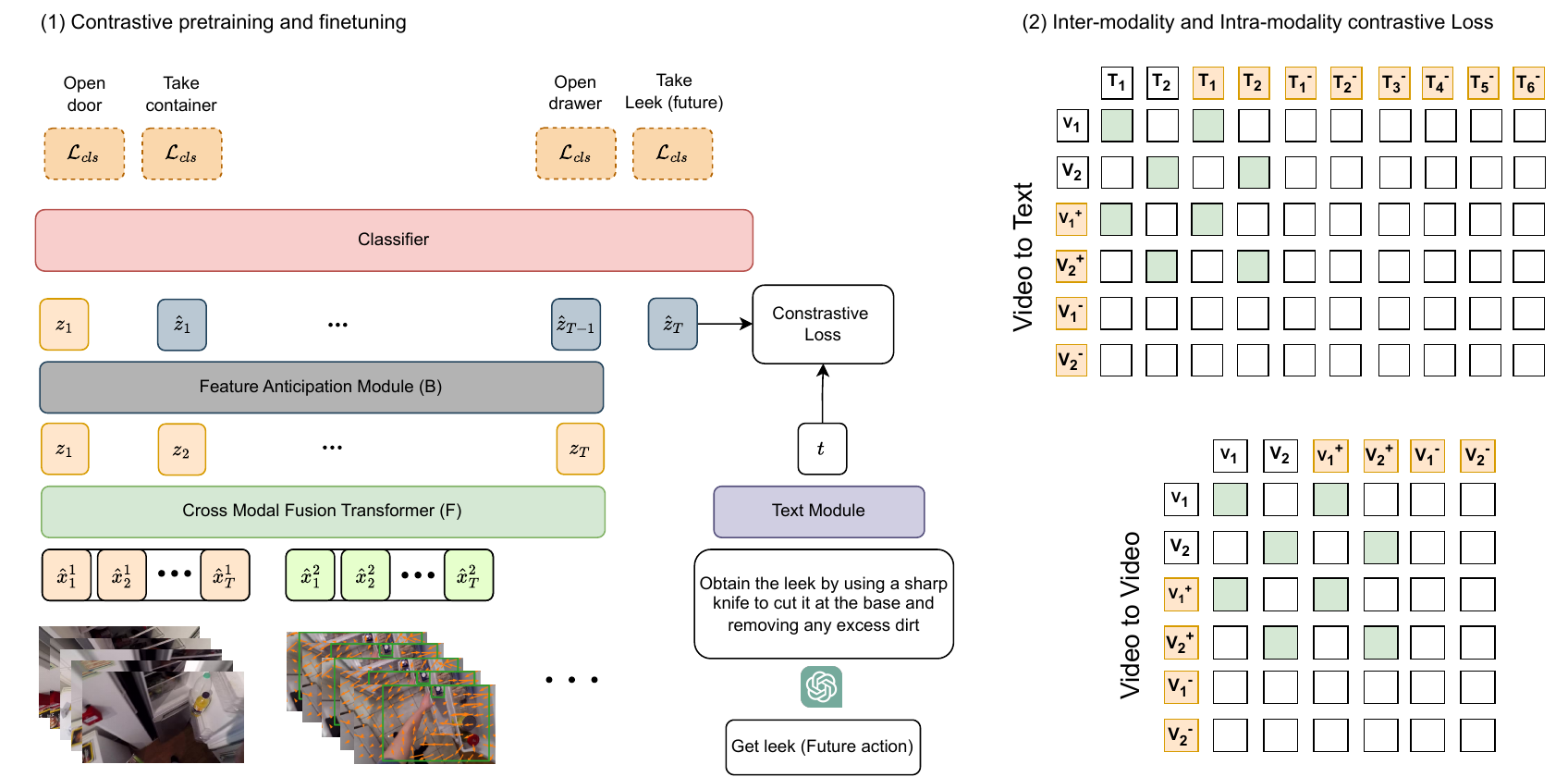}
    \caption{\textbf{Left:} 
    % Overview of our approach:  
    Our training comprises two stages: first, contrastive pre-training, where we fuse embeddings from different modalities using a Fusion Module $\mathcal{F}$, followed by an anticipation module $\mathcal{B}$.
    The output is contrasted against the rich  descriptions of future actions generated using a LLM. 
    The second stage involves fine-tuning a linear layer to predict future actions.
    \textbf{Right:} Illustration of the image-text and image-image contrastive setup.}
    \label{fig:overview}
\end{figure*}

\textbf{Language-Image Pre-training}
Training images jointly with natural language text (e.g., captions) has been established as an effective pre-training method for zero-shot learning, open vocabulary testing, and as well as classification tasks. 
CLIP \cite{radford2021learning}, ALIGN \cite{jia2021scaling}, FLorence \cite{yuan2021florence}, X-CLIP \cite{ma2022x}, UniCL \cite{yang2022unified} have shown that contrastive training on large-scale image-text pairs results in astonishing performance for zero-shot prediction.
OWL-ViT \cite{minderer2205simple} uses a CLIP-based contrastive approach to transfer image-level pre-training to open vocabulary object detection.
Similarly, CoCa \cite{yu2022coca} is not only trained on the contrastive loss, but also leverages generative modeling via a captioning loss.  
Flamingo \cite{alayrac2022flamingo} on the other hand interleaves visual data with text and produces free-form text as output, demonstrating effective performance on several downstream tasks.
Such natural language supervision also aids in video representation learning. 
For instance, \cite{bertasius2020cobe} used a visual detector to map every object instance in the video frame into its contextualized word representation obtained from narration.
Building on these works, we propose a CLIP-like contrastive pre-training approach that learns to align multi-modal features with rich descriptions of future actions.

\textbf{Multi-modal training}
Typically, modalities used for action anticipation include RGB images, optical flow, object information, IMU, and audio \cite{girdhar2021anticipative,furnari2019anticipating,sener2020temporal,wu2021learning,zatsarynna2021multi,zhong2023anticipative}. 
Features from each modality are averaged, either weighted \cite{girdhar2021anticipative} or unweighted \cite{furnari2019anticipating}, or a Multi-Layer Perceptron (MLP) is used \cite{kazakos2019epic}. 
Recently, multi-head cross attention is being employed to attend over different modalities \cite{zhong2023anticipative,liu2021cma}. 
However, training modality specific encoders can be computationally expensive.
% Instead, we explore the usage of text-based inputs as modalities, i.e., objects and actions detected in text form in lieu of visual features.
Instead, we explore the usage of objects and actions detected in text form, in lieu of visual features.
To this end, we propose an architecture which contrasts fused features from different modalities including text from actions and objects detected in the video, with descriptions generated from action labels. 
% \vspace{-1 em}
\section{Methodology}
% \vspace{-1em}
\label{sec:method}
\label{sec:ps}
% As illustrated in Figure~\ref{fig:template}, 
Given a video segment starting at $ \tau_{s}$, the goal is to anticipate future action using $\tau_{o}$ length of observed segment $\tau_{a}$ units before it, i.e., from $\tau_{s}-(\tau_{a} + \tau_{o})$ to $\tau_{s} - \tau_{a}$ as seen in Figure \ref{fig:template}.
The anticipation time $\tau_{a}$ is usually fixed for each dataset, while the observation time $\tau_{o}$ can be varied.
We extract $T$ temporally sequential inputs for $M$ modalities, denoted as $x_{i}^{m}$, $i \in \{1, \dots, T\}$ and $m \in \{1, \dots, M\}$. 

Our model architecture (shown in Figure \ref{fig:overview}) comprises two stages: contrastive pre-training and fine-tuning to perform action anticipation.
During pre-training, the model consists of $M$ modality specific feature extractors $\mathcal{B}_m, \;m \in \{1, \dots, M\}$, a fusion model $\mathcal{F}$, and an anticipative module $\mathcal{B}$. 
In the fine-tuning stage, an additional classifier is trained to predict the future action, while the rest of the model is kept frozen. 
 We utilize the fusion module from \cite{zhong2023anticipative}, and a variation of the GPT2 model used in \cite{girdhar2021anticipative} for feature anticipation, to predict $\hat{z}_{i+1}=\mathcal{D}(z_i), i \in \{1,\dots,T\}$.
 In what follows, we detail the two stages, along with the implementation details.
Throughout, all modality feature are extracted from pre-trained models. 

\subsection{Pre-training}
\label{sec:pre}
We employ a CLIP-like \cite{radford2021learning} setup, where the embeddings from different modalities (e.g., images and audio) are contrasted against text embeddings computed from text descriptions of future action classes (detailed below). 
The setup utilizes the following modality features:

\textit{Video Features:} 
Given a video segment \textit{V} consisting of \textit{T} frames, the backbone network \textit{B} extracts features for each frame. 
Following \cite{zhong2023anticipative}, we use the Swin transformer features extracted with Omnivore \cite{girdhar2022omnivore}, which was trained for action recognition.

\textit{Other Modality Features:}
For other modalities like audio, optical flow, etc., we use the features provided by the official repositories \cite{zhong2023anticipative,furnari2020rolling} (see Section \ref{sec:impl} for reference).

\textit{Text Embeddings for Descriptions of Actions and Objects:}
The embeddings for text data are extracted using a pre-trained CLIP text encoder, which is kept frozen during training, and only a modality specific projection layer is trained.
The setup for obtaining the text descriptions for actions and objects is the following: 
\textit{(i) objects in the video: } the objects present in the (current) video are detected using a pre-trained FasterRCNN model for every frame\cite{furnari2020rolling}.
They are converted into a sentence using the template: \verb|A video containing the following|  \verb|objects: <list of objects>|, and encoded using the aforementioned CLIP text encoder;
and
\textit{(ii) actions in the video:} similarly, we also generate a sentence for the (current) actions in the video using the template: 
\texttt{A video containing the following actions: <list of action>}.
As some datasets do not have dense action annotations, whenever actions are not available, we use the ``no action'' tag.
During both pre-training and action anticipation, we use ground-truth action labels. 
However, we also analyze the impact of the action recognition accuracy on action anticipation in Section \ref{sub:ablation}. 

\textit{Cross Modal Fusion:}
For fusing information from multiple modalities $x_i^m$, we use the Self-Attention Fuser (SA-Fuser) blocks from \cite{zhong2023anticipative}. 
% It applies $L$ consecutive Transformer encoders at each time step with dimentionality of $d$ and $k$ attention heads, and contains a learnable token $x^{\Lambda}$.
It contains $L$ consecutive Transformer encoders with dimentionality of $d$ and $k$ attention heads, and a learnable token $x^{\Lambda}$.
The final output is the mean of all learnable tokens.

\textit{Anticipation:}
The Fused embeddings are passed through a variation of the GPT-2 \cite{radford2019language} module to predict the future features: 
% \begin{equation}
$
    \hat{\textbf{z}}_{1}, \dots \hat{\textbf{z}}_T = \mathcal{D}(\textbf{z}_1, \dots, \textbf{z}_T)
$
% \end{equation}
where $\hat{\textbf{z}}_t$ is the predicted feature corresponding to the frame $\textbf{z}_t$ after attending to the frames $\textbf{z}_1, \dots, \textbf{z}_{t-1}$. 
We refer the reader to \cite{girdhar2021anticipative} for more details.

\label{sec:chatgpt}
\begin{figure*}[!t]
    \centering
    \includegraphics[width=1\linewidth]{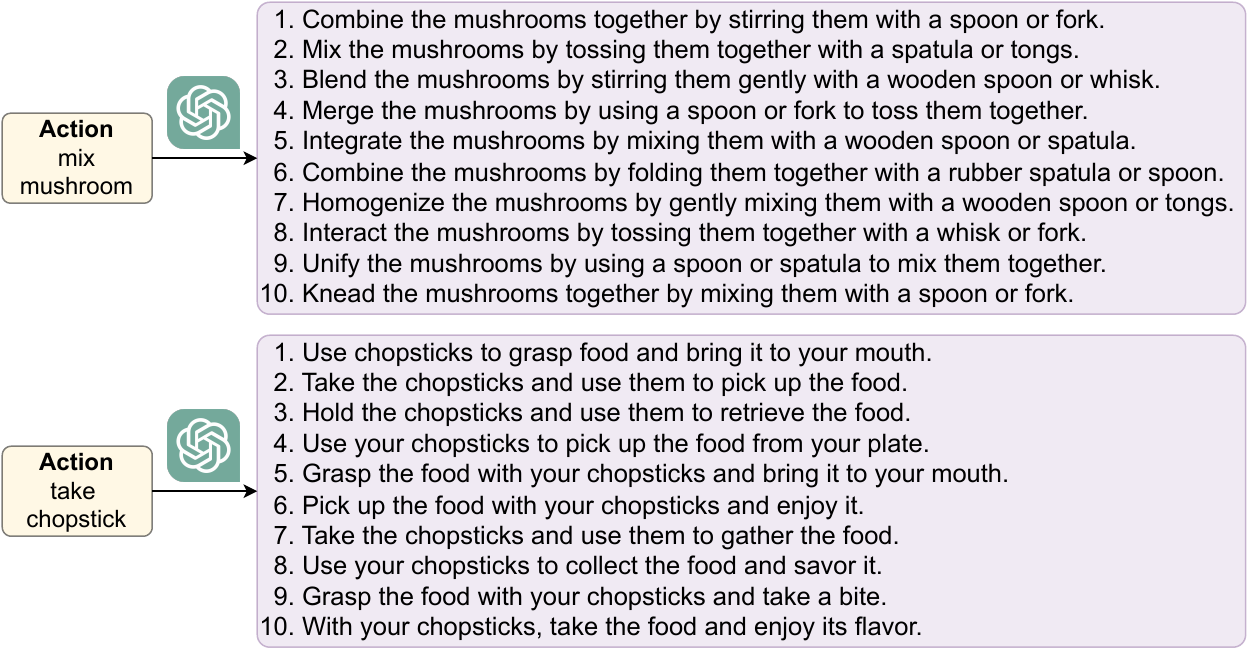}
    \caption{Descriptions generated using the ChatGPT API for actions in the EPIC-Kitchen dataset. 
    The generated descriptions add more contextual cues for the model to learn from. 
    For instance, for the action \textit{take chopsticks}, the description is already alluding to the future action of ``picking up food" or ``eating". 
    During training, we randomly select one description for every action.
    }
    % \vspace{-0.1in}
    \label{fig:chatgpt}
\end{figure*}

\textit{Text Embeddings for Rich Descriptions of Future Actions:}
we generate diverse and context rich text descriptions of the action classes using GPT3.5 \cite{brown2020language} (from the OpenAI API), by asking the system to be an expert at video based analysis and help create a caption generation system. 
We convert the class names into sentences using the prompt: \verb| Describe| \verb|<xyz> action in 1 sentence in 10| \\\verb|different ways|, and randomly select one response during training. 
We see that the descriptions generated (seen in Figure~\ref{fig:chatgpt}) are varied, and often include other objects that are interacted with for the action to take place.
% For reference, we provide examples and details about this generation in the Appendix.

\textit{Pre-training:} Features at $z_T$, which have encoded the temporal information over all observed frames, are then trained to align with the text embedding of the future action via contrastive training.
As the models were trained using smaller batch sizes, for effective contrastive learning, we augment the training with additional positive and negative samples. 
Our model is trained on features and not raw videos, and instead of applying augmentations to videos for generating additional positive samples, we follow a slow-and-fast approach.
For every sample (fast) in the batch, we create another positive sample (slow) by uniformly sampling $(1/4)T$ number of frames (denoted as $V_i^{+}$ in Figure~\ref{fig:overview}b), and the negative samples ($V_i^{-}$) are created by randomly shuffling the temporal order of the frames.
In addition, we contrast every video samples against all other action classes that do not appear in the batch ($T_i^{-}$). 
In order to limit the memory usage, we cap the number of negative text samples to 512.
So, for a batch-size of 128, we have a total of $N_v = 128 * (128*3)$ samples for the text-to-video loss, and $N_t = (128*3)*(128*2 + 512)$ samples for video-to-text loss (see Figure~\ref{fig:overview}b), an increase of $\sim$300k samples.
% In every iteration, the model has 0.0008:1 ratio of positive to negative samples, close to using a batch-size of 1024, as opposed to 0.008 when a batch-size of 128 is used. 
In every iteration, the model now sees 0.001:1 ratio of positive to negative samples instead of 0.008:1, with a batch size of 128 (for batch size = 1024, the ratio is 0.0008:1)

Similar to SLIP \cite{mu2022slip}, we also add a self-supervised learning objective.
The positive and the negative samples curated ($V_i^{+}$ and $V_i^{-}$), along with original samples ($V_i$) are trained such that similar samples are pushed closer in the embedding space.

We use standard cross entropy to train the contrastive loss. 
The loss is defined as 
\begin{equation}
\mathcal{L}_{cross} = (\mathcal{L}_{v2t} + \mathcal{L}_{t2v}) * 0.5 + \mathcal{L}_{v2v}
\end{equation}

Following AVT \cite{girdhar2021anticipative}, we also utilize self-supervised feature losses $\mathcal{L}_{feat}$ and $L_{next}$ in addition to the contrastive loss.
% In addition to the contrastive loss, following \cite{girdhar2021anticipative}, we also use the self-supervised feature loss $\mathcal{L}_{feat}$.
Therefore, our final loss function is $\mathcal{L} = \mathcal{L}_{cross} + \mathcal{L}_{feat} + \mathcal{L}_{next}$, where $\mathcal{L}_{feat}$ is defined as mean squared error between $\hat{\textbf{z}}_{t}$ and $\textbf{z}_{t+1}$, which matches the future features predicted with the true features in a self-supervised manner.

% \vspace{-0.75em}
\subsection{Action Anticipation}

Here, we fine-tune the classifier layers for the action anticipation task. 
We use the features obtained from the feature anticipation module, $\hat{\textbf{z}_T}$, in conjunction with a linear layer, and train with the cross entropy loss $\mathcal{L}_{cls}$. 
The fusion ($\mathcal{F}$) and the anticipation ($\mathcal{B}$) modules are kept frozen when the same modalities are used, and the fuser is finetuned when different modalities are used during pre-training and fune-tuning stages.

% \vspace{-0.5 em}
\begin{table}[!t]
    \centering
    \resizebox{1\columnwidth}{!}{
    \begin{tabular}{L{.15\columnwidth} L{.1\columnwidth} L{.4\columnwidth} L{.3\columnwidth}}
    \toprule
    Dataset & $\tau_a$ & Modalities & Metrics \\ 
    \midrule %\hline
    EGTEA+ & 0.5s & RGB, Flow & Top-1, cm Top-1 \\
    \hline
    EK55 & 1.0s & RGB(R), Obj(O), Flow(F), Audio(A), Objects (text)(U), Actions(text)(V) & Top-1, Top-5 \\
    \hline
    EK100 & 1.0s & RGB(R), Obj(O), Flow(F), Audio(A), Objects (text)(U), Actions(text)(V) & Recall@5 \\ 
    \bottomrule %\hline
    \end{tabular}
    }
    \caption{Modalities and metrics used for different datasets.}
    \vspace{-1em}
    \label{tab:modalities}
\end{table}

\subsection{Implementation details}
\label{sec:impl}
We process the input videos similar to \cite{girdhar2021anticipative}, and sample 16 frames at 1 fps, by setting
% the observed time 
$\tau_o=16s$. 
We use Swin Transformer based RGB features provided by \cite{zhong2023anticipative}, which were extracted from the Omnivore \cite{girdhar2022omnivore} network, originally trained for action recognition.
We use the pre-trained CLIP text encoder, processor, and tokenizer, provided by \cite{wolf2020transformers} for processing all text inputs. 
During pre-training, the encoded features are projected to 1024 dimensions, before passing through the fusion and the anticipative modules. 
In the fine-tuning stage, the fused features are classified using a single linear layer.
For both stages, we use the SGD$+$momentum optimizer, using a learning rate $1e^{-3}$ and weight decay $1e^{-6}$ for 50 epochs.
Further, we employ a cosine annealing learning rate schedule with a warmup for 20 epochs, and the training is performed on a single Nvidia A40 GPU, with a batchsize of 128.
For the optical flow and object features, we use the official RULSTM \cite{furnari2020rolling} repository, and for audio, we use features provided by \cite{zhong2023anticipative}
\footnote{Please refer to \cite{zhong2023anticipative} for details about the feature extraction for different modalities.}.
Our code, weights, and the action descriptions generated will be publicly released upon acceptance.

\section{Experiments}
\label{sec:experiments}
\subsection{Experimental setup}
\label{sec:setup}
\textbf{\textit{Datasets and metrics:} }
We evaluate on three action anticipation datasets: \emph{(i)} EpicKitchens100 (EK100) ~\cite{damen2020rescaling}, reporting the class-mean Recall@5 for actions, verbs and nouns;
\emph{(ii)} EpicKitchens55 (EK55)~\cite{damen2018scaling}, where we report the Top-1 and Top-5 for actions, verbs and nouns, through standard train and val splits; and
\emph{(iii)} EGTEA Gaze+ ~\cite{li2018eye}, in which we report the performance on the first split of the dataset at $\tau_a$ = 0.5s, and the metrics include Top-1 and class-mean(cm) Top-1 accuracies for actions, nouns and verbs.
We add further details about these datasets in the Appendix.

\noindent\textbf{\textit{Modalities:}}
We summarize the modalities and metrics used in Table~\ref{tab:modalities}. 
We use pre-trained TSN weights provided by the official repositories~\cite{furnari2020rolling,zhong2023anticipative} for object features, audio, and flow.
We use the objects detected using a FasterRCNN model trained on Epic-Kitchen 55 dataset~\cite{furnari2020rolling}, with a threshold of 0.15 and pick the top 5 objects for every frame in the video.
For actions, we use labels provided in the dataset during training and evaluation. 
We evaluate the impact of action recognition accuracy and discuss the results in Section \ref{sub:ablation}.

\noindent\textbf{\textit{Baselines:}}
We evaluate our approach against the state-of-the-art for action anticipation, including,  RULSTM ~\cite{furnari2020rolling}, AVT~\cite{girdhar2021anticipative}, ActionBanks~\cite{sener2020temporal}, AFFT~\cite{zhong2023anticipative}, and MeMViT~\cite{wu2022memsit}.
% \footnote{Please see the Appendix for details about the baselines.}.
RULSTM~\cite{furnari2020rolling} leverages a `rolling' LSTM to encoder the past and an `unrolling' LSTM to predict the future.
ActionBanks~\cite{sener2020temporal} improves over RULSTM by carefully leveraging long-term action blocks and non-local blocks.
AVT~\cite{girdhar2021anticipative} uses an attention-based video modelling architecture that attends to previous frames to anticipate the future.
MeMViT~\cite{wu2022memsit}, on the other hand, processes videos online by using cache ``memory", through which the model learns to refer prior context for long-term anticipation.
AFFT~\cite{zhong2023anticipative} improves on AVT by using multiple modalities, and using self-attention modules to fuse the features together.
We re-train the AFFT model on our local environment setup for fair comparison, and observe a small discrepancy in performance relative to the published paper. 
As the goal of this paper is to demonstrate the effectiveness of learning from text embeddings, we do not compare against other state-of-the-art methods that have a substantially different architectures like ~\cite{roy2021action,roy2023predicting,zhao2023antgpt}.
For example, AntGPT~\cite{zhao2023antgpt} introduces a promising alternate way of predicting future actions by fine-tuning LLMs.
Yet, we do not compare against it, as we do not use LLMs to infer our outputs or predict goals and actions, rather only use it to generate detailed descriptions.
Additionally, it reports performance in few-shot settings for the Ego4D dataset (which we did not evaluate on) making one-to-one comparison hard. 

\vspace{-0.1em}
\noindent\textbf{\textit{ChatGPT generated action descriptions:}}
\label{sec:gpt_actions}
We provide examples of the descriptions generated for actions using the ChatGPT API (with  GPT3.5 Turbo) in Figure~\ref{fig:chatgpt}.
In the descriptions, there are generally mentions of other objects that are used when the action takes place.
For example, the text descriptions for ``take chopsticks'' are \emph{``Use chopsticks to grasp food and bring it to your mouth''}, \emph{``Take the chopsticks and use them to pick up the food''}, etc., giving context about other objects in contact with hand. 
Similarly, descriptions for the ``mix mushroom'' action often involve words such as tongs, spoons or a spatula.

\begin{table}[!t]
% \centering
\def\arraystretch{1.2}
\resizebox{1\columnwidth}{!}{
\begin{tabular}{l c c c c c c c}
\toprule
\multicolumn{1}{c}{Model} & \multicolumn{3}{c}{Top-1} &  & \multicolumn{3}{c}{Class mean acc} \\ \cline{2-4} \cline{6-8} 
 & Verb & Noun & Act. &  & Verb & Noun & Act. \\ 
\midrule 
I3D-Res50~\cite{carreira2017quo} & 48.0 & 42.1 & 34.8 &  & 31.3 & 30.0 & 23.2 \\
FHOI~\cite{liu2020forecasting} & 49.0 & 45.5 & 36.6 &  & 32.5 & 32.7 & 25.3 \\
AVT(TSN)~\cite{girdhar2021anticipative} & 51.7 & 50.3 & 39.8 &  & \textbf{41.2} & 41.4 & 28.3 \\
% AFFT~\cite{zhong2023anticipative} & 53.4 & 50.4 & 42.5 &  & 42.4 & 44.5 & 35.2 \\ %\hline
AFFT~\cite{zhong2023anticipative} & \textbf{52.1} & \textbf{50.7} & \textbf{41.4} &  & 38.4 & \textbf{43.7} & \textbf{31.8} \\ %\hline
\midrule 
Ours (R $\rightarrow$ RF) & {51.4}& {49.7} & { 40.8} &  & {38.9} & {43.3} &{31.3} \\ %\hline
\bottomrule
\end{tabular}
}
\caption{
\textbf{EGTEA Gaze+: } Model performance for Split=1 at $\tau_{a}=0.5s$. \textbf{Bolded} values indicate highest score, and $\rightarrow$ denotes the modalities used for pre-training and fine-tuning.}
\label{Tab:egtea}
\end{table}

\begin{table}[!t]
% \centering
\def\arraystretch{1.2}
\centering
\resizebox{1\columnwidth}{!}{
\begin{tabular}{cccccccccc}
\hline
Method & \multicolumn{2}{c}{Verb} &  & \multicolumn{2}{c}{Noun} &  & \multicolumn{2}{c}{Action} &  \\ \cline{2-3} \cline{5-6} \cline{8-9} 
 & Top-1 & Top-5 &  & Top-1 & Top-5 &  & Top-1 & Top-5 \\ \hline
RULSTM & 32.4 & 79.6 &  & 23.5 & 51.8 &  & 15.3 & 35.3 \\
ActionBanks & \textbf{35.8} & {80.0} &  & 23.4 & 52.8 &  & 15.1 & 35.6 \\
AVT & - & - & & - & - & & 14.4 & 31.7 \\
AVT+ & 32.5 & 79.9 &  & 24.4 & 54 &  & 16.6 & {37.6} \\
AFFT & 34.9 & 78.7 &  & {26.2} & 53.9 &  & 17.0 & 34.3 \\ \hline
Ours (R $\rightarrow$ R) & 32.4 & 80.1 &  & 28 & 56.4 &  & 16 & 36.5 \\
Ours(ROFA $\rightarrow$ ROFA) & 33 & 79.4 &  & 26 & 55.5 &  & 14.9 & 35.9 \\
% \rowcolor{green}
Ours(R $\rightarrow$ ROFA) & 32.5 & 80.4 &  & 27.8 & 57 &  & 16.5 & 38.1 \\
Ours* (R $\rightarrow$ ROFA+UV) & {34.3} & \textbf{80.6} && \textbf{29.7} & \textbf{58.8} && \textbf{17.9} & \textbf{39.8} & \\ 
\hline
\end{tabular}
}
\caption{\textbf{EK55:} Comparison of state-of-the-art methods on the validation set of EK55 using the modalities (ROFA). 
* indicates that additional action (V) and objects (U) information was provided in the text form. 
% w/o gpt indicates that the model was not pre-trained using GPT descriptions. 
R indicates that RGB features were used, ROFA refers to RGB, Obj(TSN), Flow (TSN) and Audio features.}
\label{tab:ek55}
\end{table}

\subsection{Comparison Against Baselines}
\textbf{EGTEA+}
In Table~\ref{Tab:egtea}, we compare our results on split 1 (as in ~\cite{liu2020forecasting}) at $\tau_a=0.5s$. 
In addition to the RGB data, we use the flow data provided by ~\cite{furnari2020rolling}.
Similar to AFFT, we use the pre-trained TSN features.
We also note that the results for AFFT were obtained by using the official code on our local environment.
We observe that our approach does not improve performance on EGTEA+, in contrast to other larger datasets.
The smaller scale of EGTEA+ is not a good match for contrastive learning, which is generally sensitive to data size and sample variety, and thereby, does not result in performance improvement.

\noindent\textbf{Epic-Kitchens}
In Table~\ref{tab:ek55} and Table~\ref{tab:EK100}, we compare the performance of our method to the state-of-the-art for the EK55 and EK100 datasets. 
For EK55, we obtain the results for the AFFT baseline using the authors' code.
First, we consider the performance of our approach when trained using only the (single) RGB modality. 
As the fusion module is a block of transformer layers, they act as feature encoder layers when there no  modalities to fuse.  
We observe that our method has a 2\% absolute improvement over AFFT, a multi-modal baseline, and $\sim$5\% to AVT's single modality performance.
When the model is pre-trained and fine-tuned with multiple modalities (ROFA$\rightarrow$ROFA), it outperforms AFFT for the Top-5 metrics, yet performs poorly compared to pre-training solely on RGB. 
We believe that during pre-training, the model can find it challenging to fuse all the modality features and then align them with text embeddings.
Additionally, we hypothesize that an Imagebind-like training setup might be beneficial, however, we do not evaluate this scenario as ImageBind trains modality specific encoders, whereas we do not.
This leads us to evaluate pre-training with just the RGB modality, while fine-tuning the fusion and classifier layers with multiple modalities (R$\rightarrow$ROFA). 
With this training strategy, we see an improvement of ~2\% compared to the (ROFA$\rightarrow$ROFA) training,  outperforming our single modality performance by 0.5\% for Top-1 and 1.3\% for Top-5 metric for actions.
Adding additional information about the objects and actions, we see an absolute improvement of 1\% for Top1, and 5\% for Top5.

\begin{table}[!t]
\centering
  \setlength{\tabcolsep}{6pt} % Default value: 6pt
    \def\arraystretch{1.2}
    \resizebox{1\columnwidth}{!}{
    \begin{tabular}{cccccccccccc}
    \hline
    Method & \multicolumn{3}{c}{Overall} &  & \multicolumn{3}{c}{Unseen} &  & \multicolumn{3}{c}{Tail} \\ \cline{2-4} \cline{6-8} \cline{10-12} 
     & Verb & Noun & Action &  & Verb & Noun & Action &  & Verb & Noun & Action \\ \hline
    RULSTM & 27.8 & 30.8 & 14.0 &  & 28.8 & 27.2 & 14.2 &  & 19.8 & 22.0 & 11.1 \\
    TempAgg & 23.2 & 31.4 & 14.7 &  & 28 & 26.2 & 14.5 &  & 14.5 & 22.5 & 11.8 \\
    AVT & 30.2 & 31.7 & 14.9 &  &-&-&-&  &-&-&-\\
    AVT+ & 28.2 & 32.0 & 15.9 &  & 29.5 & 23.9 & 11.9 &  & 21.2 & 25.8 & 14.1 \\
    MeMViT & \textbf{32.3} & {37.0} & 17.7 &  & 28.6 & 27.4 & 15.2 &  & {25.3} & 31.0 & 15.5 \\
    AFFT(Swin+) & 22.8 & 34.6 & 18.5 &  & 24.8 & 26.4 & 15.5 & & 15.0 & 27.7 & 16.2\\ 
    AFFT (re) & 22.4 & 32.4 & 18.1 &  & 26.5 & 26.8 & 15.3 & & 14.6 & 24.3 & 15.9\\ \hline
    Ours (R $\rightarrow$ R) & 30.1 & 32 & 16 & & 32.7 & 28.4 & 15.3 & & 23.4 & 25.3 & 13.8 \\
    Ours (R $\rightarrow$ ROFA) & 31.9 & 35.9 & 17.3 & & 32.5& 30.2 & 14.5 & & \textbf{25.9} & 30.3 & 15.4\\
    Ours$^*$ (R $\rightarrow$ ROFA+UV) & {31.3} & \textbf{47.8} & \textbf{23.8} &  & \textbf{34.5} & \textbf{42.8} & \textbf{24} &  & {23.8} & \textbf{41.9} & \textbf{20.3} \\ \hline
    \end{tabular}
}
\caption{\textbf{EK100}: comparison of state-of-the-art method on the validation set of EK100 using modalities provided by \cite{furnari2020rolling}. 
MeMViT uses only RGB data, while the rest use multiple modalities. 
R indicates that only RGB features were used, ROFA refers to RGB, Obj(TSN), flow(TSN) and Audio features.
* indicates that additional action (V) and object(U) modalities in the text form were used.
% Every method is indicated with (pre $\rightarrow$ fine-tuning) setting.
\textbf{Bolded} values indicate the best performing method, and $\rightarrow$ denotes the modalities used for pre-training and fine-tuning.}
\label{tab:EK100}
\end{table}

For EK100, we compare our two-stage network against single-stage methods, and with our method of using action and object information in the text form in Table \ref{tab:EK100}. 
Similar to EK55, we see that our method trained only on RGB outperforms AVT, while performing comparably to MeMViT. 
However, MeMViT is a method that is directly trained on the videos, while we used pre-extracted features.
Using multiple modalities (R$\rightarrow$ROFA) performs similarly to AFFT for actions, but shows a signification improvement in verb and noun predictions. 
In addition, with action and object information, we see a clear improvement across all predictions, particularly in the unseen and the tail category.
This indicates that the model has effectively learned from the additional context provided by the text representations.

\begin{table}[t]
\centering
  \setlength{\tabcolsep}{6pt} % Default value: 6pt
\def\arraystretch{1.2}
\resizebox{1\columnwidth}{!}{
\begin{tabular}{ccccccccccc}
\hline
Method & \multicolumn{2}{c}{Verb} &  & \multicolumn{2}{c}{Noun} &  & \multicolumn{2}{c}{Action} &  \\ \cline{2-3} \cline{5-6} \cline{8-10} 
 & Top-1 & Top-5 &  & Top-1 & Top-5 &  & Top-1 & Top-5 & Recall@5 \\ \hline
Ours (w/ gpt) & \textbf{32.8} & 79.8 &  & 27.8 & 56.5 &  & 15.6 & \textbf{36.8} & 16.1 \\
Ours (w/o gpt) & 31.9 & 79.6 &  & 26.6 & 56.1 &  & 14.8 & 36.7 & 16 \\
Ours (w/o Aug) & 31.9 & 79.5 &  & 26.6 & 56.1 &  & 14.8 & 36.3 & 14.8 \\
Ours (w/ $L_{v2v}$) & 32.4 & \textbf{80.1} &  & \textbf{28} & \textbf{56.4} &  & \textbf{16} & 36.5 & \textbf{17.5}\\
\hline
\end{tabular}
}
\caption{\textbf{EK55 Ablations}: comparing different training losses and protocols on the validation set of EK55 using only RGB.
w/ and w/o gpt indicate pre-training with/without using descriptions of future actions. 
w/o Aug indicates that slow-fast and negative samples were not appended to the batch samples during the pre-training, while all other methods were not trained using $L_{v2v}$ loss, w/ $L_{v2v}$ is trained with the loss in addition to other losses and data augmentations.}
\label{tab:ek55_ablation}
\end{table}

\begin{figure*}[t]
    \centering
    \includegraphics[width=0.75\linewidth]{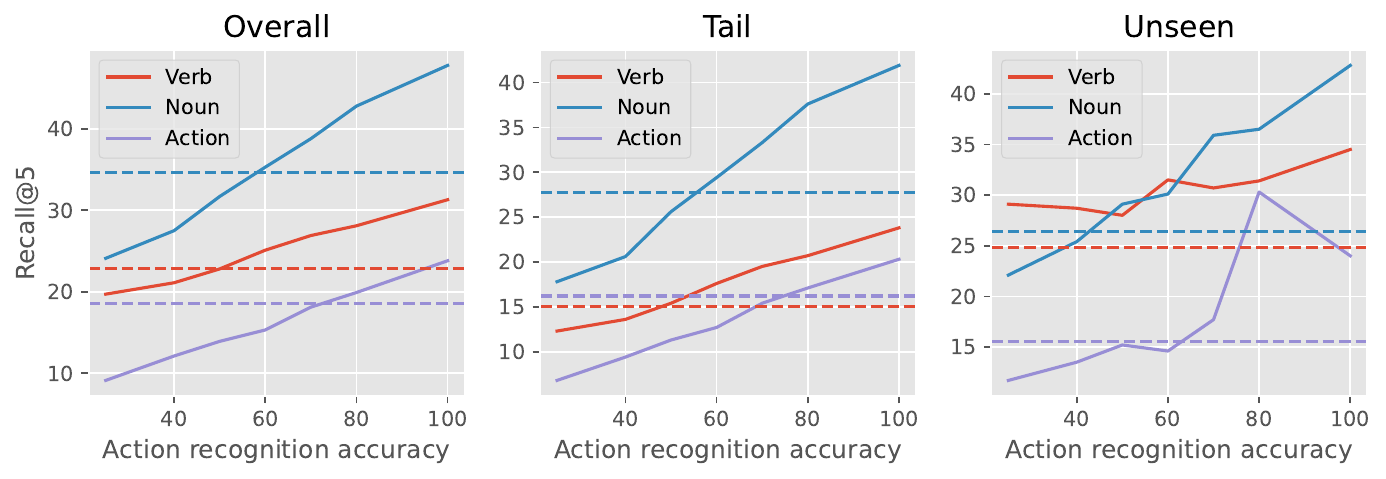}
    \caption{Impact of action recognition accuracy on the prediction of \textcolor{orange}{verbs}, \textcolor{cyan}{nouns} and \textcolor{purple}{actions} for EK100. Values in dashed lines are the corresponding results from the AFFT baseline. 
    }
    \label{fig:actions}
\end{figure*}

\begin{table}[t]
\centering
\def\arraystretch{1.2}
\resizebox{1\columnwidth}{!}{
\begin{tabular}{cccccccccccc}
    \hline
Method & \multicolumn{3}{c}{Overall} &  & \multicolumn{3}{c}{Unseen} &  & \multicolumn{3}{c}{Tail} \\ \cline{2-4} \cline{6-8} \cline{10-12} 
 & Verb & Noun & Action &  & Verb & Noun & Action &  & Verb & Noun & Action \\ \hline
Ours (ROFA$\rightarrow$ROFA) & 28& 33.9 & 15.8 & & 29.4 & 28 & 16.6 & & 20.5 & 27 & 13.3 \\
Ours (R $\rightarrow$ R) & 30.1 & 32 & 16 & & 32.7 & 28.4 & 15.3 & & 23.4 & 25.3 & 13.8 \\
Ours (R $\rightarrow$ RV) & 28.1 & 44.3 & 21.8 & & 36.7 & 41.5 & 23.3 & & 20.0 & 37.9 & 18.6  \\
Ours (R $\rightarrow$ RAV) & 30.6 & 44.6 & 22.4 &  & \textbf{41.0} & 40.5 & 21.9 & & 23.1 & 38.4 & 19.2 \\
Ours (R $\rightarrow$ ROFA+UV) & \textbf{31.3} & \textbf{47.8} & \textbf{23.8} &  & {34.5} & \textbf{42.8} & \textbf{24} &  & \textbf{23.8} & \textbf{41.9} & \textbf{20.3} \\ \hline
\end{tabular}
}
% \caption{\textbf{Ek100} val sets. Impact of different modalities\apoorva{Not complete}}
\caption{\textbf{EK100 Ablations}: Impact of different modalities on model performance.}
\vspace{-1em}
\label{tab:ek100_ablation}
\end{table}

\subsection{Ablations and Analysis}
\label{sub:ablation}
\noindent\textbf{Impact of modalities:}
In Table~\ref{tab:ek100_ablation}, we explore the contributions of various modalities to our performance on EK100. 
Using RGB as one of the modalities, we examine the contributions by audio and actions to the performance. 
In detail, the modality contributions are discussed in the Appendix.
We see that action and audio provide complementary information that RGB alone does not, leading to better performance. 

\noindent\textbf{Impact of different training settings:}
We evaluate the effect produced by the losses and data augmentations used during pre-training in Table ~\ref{tab:ek55_ablation}, where: 
\textit{(i) w/ gpt} indicates that ChatGPT generated action descriptions were used during pre-training (as detailed in Section~\ref{sec:setup});
\textit{(ii) w/o gpt} involves pre-training with the simple template - \verb|This is a| \verb|video clip with action| \verb|<xyz>|;
\textit{(iii) w/o Aug} indicates that during training, the batch samples were not appended with positive and negative samples from the slow-fast and randomly shuffled features(detailed in Section~\ref{sec:pre});
and 
\textit{(iv) w/ $L_{v2v}$} contains the self-supervised loss in addition to other losses  during pre-training. 
We observe that using the richer descriptions from ChatGPT and the self-supervised loss $L_{v2v}$ boosts the action prediction performance by 1.2\% for Top-1, 0.2\% for Top-5, and 3\% for recall@5, indicating their necessity during training.

\noindent\textbf{Effect of the Accuracy of Actions: }
In Figure~\ref{fig:actions}, we evaluate the impact of having access to accurate actions on action anticipation.
For this evaluation, we vary the accuracy \%-age of ground-truth action labels used.
% by randomly sampling actions for every frame. 
Therefore, when the accuracy of actions is 20\%, it indicates that 80\% of actions during training are incorrect (i.e., they are randomly sampled). 
We notice that as the action recognition accuracy increases, the noun prediction performance also increases drastically. 
% Adding recognized actions as an additional modality starts to aid in performance when the accuracy of action recognition exceeds 70\%. 
When the action recognition accuracy increases to 70\%, we see that our method starts outperforming the AFFT baseline.
% \harish{confusing sentence I think}
However, for unseen classes, an action recognition accuracy of 55\% results in performance increase.
This observation also supports that accurate action recognition is needed for accurate action anticipation.
Overall, we observe that with accurate action and object recognition systems, inputs in the text format can greatly improve prediction performance, without having to train modality specific encoders. 

\begin{table}[t]
\centering
\renewcommand{\arraystretch}{1.2}
\resizebox{1\columnwidth}{!}{
\begin{tabular}{ccccccccccc}
\hline
Method & \multicolumn{2}{c}{Verb} &  & \multicolumn{2}{c}{Noun} &  & \multicolumn{3}{c}{Action} &  \\ \cline{2-3} \cline{5-6} \cline{8-10} 
 & Top-1 & Top-5 &  & Top-1 & Top-5 &  & Top-1 & Top-5 & Recall@5 \\ \hline
Ours (baseline) & \textbf{32.8} & 79.8 &  & \textbf{27.8} & 56.5 &  & \textbf{15.6} & 36.8 & 16.1 \\
Ours (GPT corrected) & 28.4 & \textbf{79.9} &  & 25.2 & \textbf{57.1} &  & 13 & \textbf{37.6} & \textbf{16.5} \\
\hline
\end{tabular}
}
\caption{EK55: Accuracy after using GPT-4 to correct the predictions.}
\vspace{-1em}
\label{tab:ek55_ablation_supp}
\end{table}

\begin{figure*}[!t]
    \centering
    \includegraphics[width=0.75\linewidth]{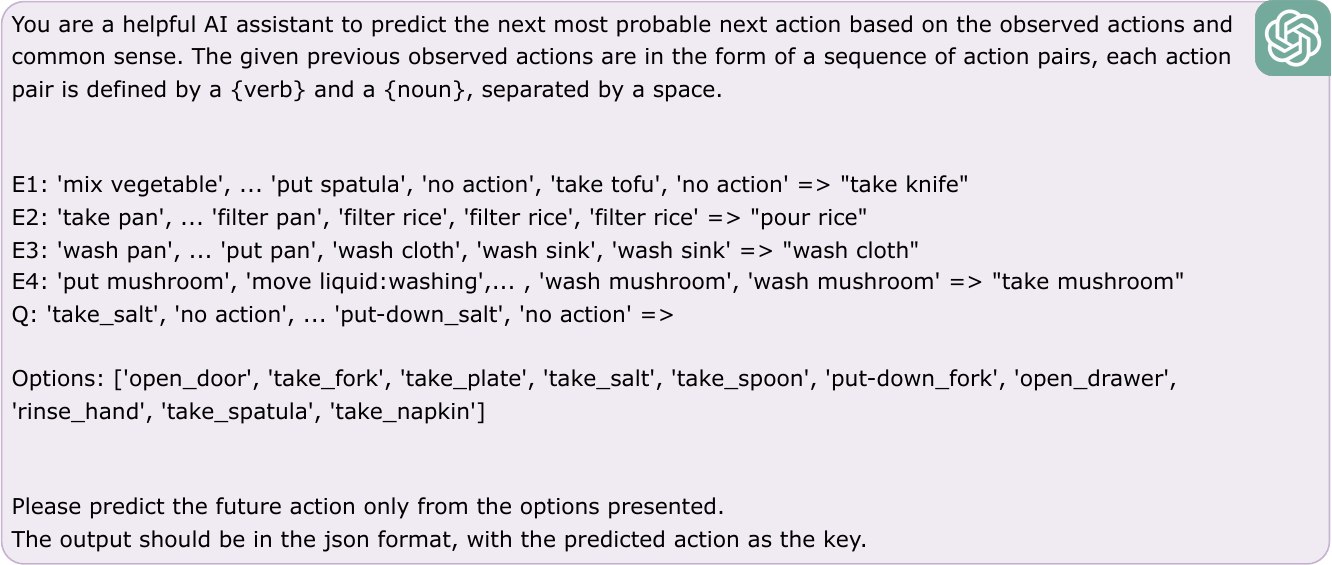}
    \caption{Prompt provided to ChatGPT to pick most plausible future action.}
    \label{fig:prompt}
\end{figure*}

\begin{table*}[h]
\centering
\small
\resizebox{0.97\linewidth}{!}{
\begin{tabular}{P{8cm} P{2cm} P{5.0cm} P{2cm}}
\toprule
Observed Actions & Future Action (corrected) & Future Actions (predicted) & GT \\
\midrule
no action, no action, open\_door, no action, take\_container, take\_container, no action, take\_lid, take\_lid, no action, no action, no action, no action, take\_lid, take\_lid, take\_lid & \textcolor{green1}{put lid} & 
\textcolor{red}{close\_door}, take\_pan, \textcolor{green1}{put lid}, put-down\_pan, open\_door, put-down\_box:cereal, take\_box:fruit, take\_colander, put-down\_colander, take\_bag:cereal
% \textcolor{red}{close\_door}, take\_pan, \textcolor{green1}{put lid}, put-down\_pan, open\_door 
& put lid \\
\midrule
close\_fridge, close\_fridge, no action, open\_bag:cereal, open\_bag:cereal, open\_bag:cereal, open\_bag:cereal, open\_bag:cereal, open\_bag:cereal, open\_bag:cereal, open\_bag:cereal, open\_bag:cereal, open\_bag:cereal, open\_bag:cereal, open\_bag:cereal, open\_bag:cereal & 
\textcolor{red}{take\_bowl} & 
fold\_bag:rice, put-down\_bag, close\_bag:rice, \textcolor{green1}{open\_bag:cereal}, take\_bag:cereal, place\_salad, put\_packet:crisp, get\_salad, take\_bowl, put-in\_bag
% fold\_bag:rice, put-down\_bag, close\_bag:rice, \textcolor{green1}{open\_bag:cereal}, take\_bag:cereal 
& open\_bag:cereal \\
\midrule
stir\_spatula, stir\_spatula, put-down\_spatula, open\_container, open\_container, take\_onion, take\_onion, take\_onion, take\_onion, close\_container, close\_container, close\_container, no action, no action, take\_spatula, take\_knife
&
\textcolor{green1}{cut\_onion}
&
\textcolor{red}{put-down\_spatula, take\_spatula, open\_container, put-down\_onion, put-down\_knife,} put\_container, \textcolor{green1}{cut\_onion}, take\_container, close\_container, take\_onion
&
cut\_onion \\
% fold\_filter, fold\_filter, fold\_filter, fold\_filter, fold\_filter, fold\_filter, put\_filter:water, put\_filter:water, put\_filter:water, put\_filter:water, no action, no action, take\_coffee, take\_coffee, take\_coffee, no action & \textcolor{red}{put-down\_bag} & 
% % \textcolor{red}{take\_bread, put-down\_bread, open\_bag:cereal, fold\_bag:rice, open\_door}
% \textcolor{red}{take\_bread, put-down\_bread, open\_bag:cereal, fold\_bag:rice, open\_door}, put-down\_box:cereal, close\_bread, put-down\_bag, put\_packet:crisp, \textcolor{green1}{open\_fridge}
% & open\_fridge \\
\midrule
take\_salt, no action, no action, no action, put-down\_salt, put-down\_salt, put-down\_salt, put-down\_salt, put-down\_salt, put-down\_salt, no action, no action, no action, no action, no action, no action & \textcolor{red}{take\_spoon} & 
% \textcolor{green1}{open\_door}, take\_fork, take\_plate, take\_salt, take\_spoon, put-down\_fork 
\textcolor{green1}{open\_door}, take\_fork, take\_plate, take\_salt, take\_spoon, put-down\_fork, open\_drawer, rinse\_hand, take\_spatula, take\_napkin
& open\_door \\
\bottomrule
\end{tabular}
}
\caption{Examples showing the past observed actions, GPT4 corrected action, predicted actions by our model, and ground-truth.
Actions in the ``Future Actions (predicted)'' are in descending order of probability.}
\label{tab:gpt_examples}
\end{table*}

\noindent\textbf{Using GPT-4 to refine predictions}
For EK55, we also explore using ChatGPT (GPT-4) to reason about the future action, given a sample set of examples from the train set, and a list of actions to choose from. 
We provide the Top-10 actions predicted by our model, and ask ChatGPT to pick the most likely action, given the actions in the observed timeframe (see Figure \ref{fig:prompt} for reference).
The previous observed actions are the ground truth, and the options are the top-10 results generated by the model using the ROFA modalities.
We provide examples in Table~\ref{tab:gpt_examples}.

From Table~\ref{tab:ek55_ablation_supp}, we observe that while the Top-5 and recall performance improve for Verb, Noun and Actions, the Top-1 performance drops for all predictions. 
As we see in Table~\ref{tab:gpt_examples}, while GPT-4 can refine the predictions in some cases, it also produces incorrect answers especially in cases where several ``no action'' are present (as the predictions are solely based on the text input of the past actions). 
When the correct action is not present in the Top-5, but is in Top-10, GPT-4 is capable of refining the prediction.
It should be noted when the correct action is not present in the Top-10 predictions at all, GPT-4 is not capable of predicting the right action, as we force it to pick one of the 10 actions.

\section{Conclusion and Future Work}
In this work, we presented Multi-Modal Contrastive Anticipative Transformer(M-CAT), a video transformer-based approach for predictive action anticipation.
We developed a two-stage process: first, contrastive pre-training between fused features from multiple modalities and rich descriptions of future actions, encoded through a text encoder; and 
second, fine-tuning, where the classifier (and fusion layers) are updated while predicting the future action. 
We evaluated and observed that object and action descriptions, added through simple text templates, can substantially improve anticipation performance.
In addition, the use of richer descriptions of future actions for contrastive pre-training was beneficial.  
We also analyzed the effect of different modalities on performance, and the impact of the accurate actions on anticipation.
In the future, we aim to utilize a pre-training stage similar to ImageBind~\cite{girdhar2023imagebind}, which learns across multiple modalities and datasets.
%%%%%%%%% REFERENCES

% \newpage
\vspace{2in}
\appendix
\begin{figure*}[h]
    \centering
    \includegraphics[width=0.85\linewidth]{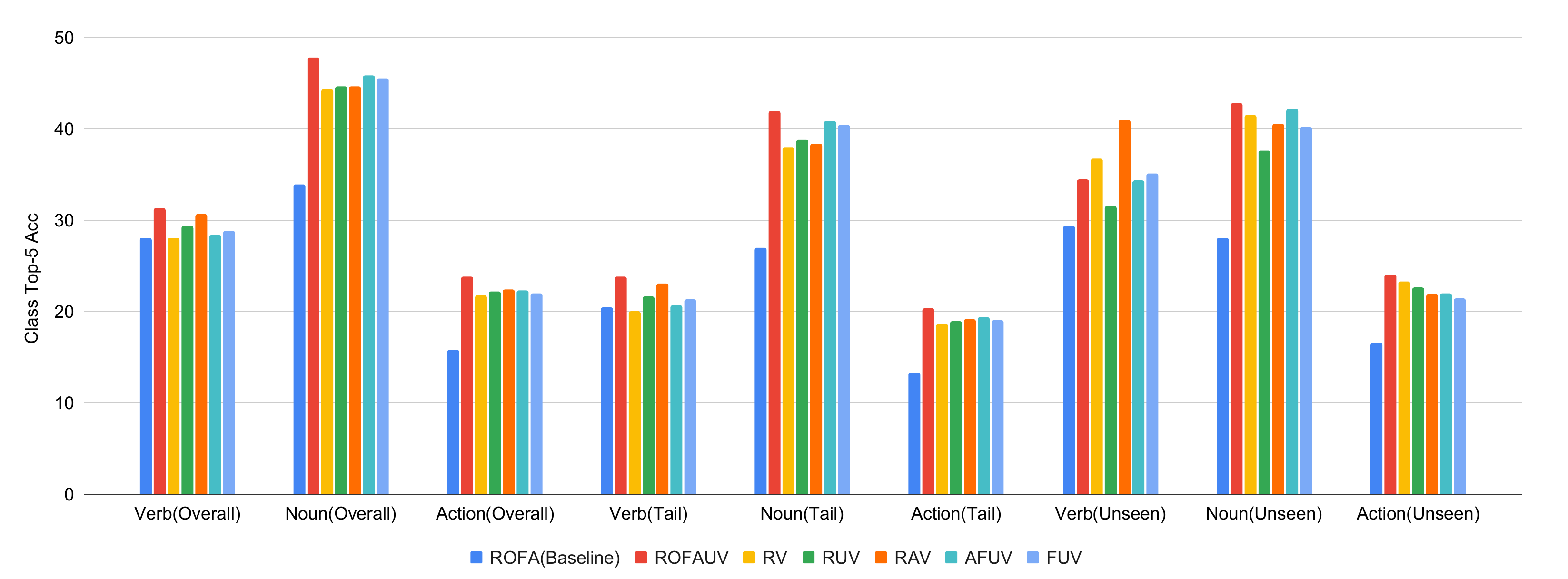}
    \caption{
    Recall@5 for verb, noun and actions on the EK100 dataset for different modality combinations. 
    The \textcolor{bluee}{first bar} is the baseline (i.e., AFFT) using (ROFA) modalities. 
    Objects and actions are used as input by converting them to text through -- ``A video containing the objects/actions $<$xyz$>$", and embeddings from the text encoder are used in the fusion module.
    ROFAUV stands for - R(RGB), O(Obj features), F(Flow), A(Audio), U(FasterRCNN detected objects in text form), V(Actions in text form) modalities}
    \label{fig:modalities}
\end{figure*}

\section{Datasets}
\textbf{\textit{Datasets and metrics:} }
We evaluate our approach on three popular action anticipation datasets:
\emph{(i)} \textit{Epic-Kitchens 100 (EK100)} ~\cite{damen2020rescaling}, which is a large egocentric video dataset with 700 long unscripted videos of cooking activities totaling  100 hours.
The dataset consists of 90K segments, and has 3807 action classes, 97 verbs and 300 nouns. 
We report the class-mean Recall@5 for actions, verbs and nouns;
\emph{(ii)} \textit{EpicKitchens 55 (EK55)}~\cite{damen2018scaling} is an earlier version of Epic-Kitchens 100.
For comparison to existing approaches, we report the validation accuracy on this dataset as well. 
EK55 has about 39K segments, and 2513 action classes, 124 verbs and 351 noun classes. 
For EK55, we report Top-1 and Top-5 for actions, verbs and nouns. 
We use the standard train and val splits to report performance. 
\emph{(iii)} \textit{EGTEA Gaze+} ~\cite{li2018eye}, an egocentric dataset containing about 10K segments, and 19 verbs, 51 nouns and 106 unique actions. 
Following \cite{girdhar2021anticipative}, we report the performance on the first split of the dataset at $\tau_a$ = 0.5s. 
We report the Top-1 and class-mean(cm) Top-1 accuracies for actions, nouns and verb.

% \section{Baselines}
% In addition to comparing our method to its variants containing different modalities, we also evaluate against the state-of-the-art for action anticipation, including:  RULSTM ~\cite{furnari2020rolling}, AVT~\cite{girdhar2021anticipative}, ActionBanks~\cite{sener2020temporal}, AFFT~\cite{zhong2023anticipative}, and MeMViT~\cite{wu2022memsit}.
% % \harish{might want to consider mentioning briefly what the approaches are -- probably in the RW?}
% RULSTM~\cite{furnari2020rolling} leverages a `rolling' LSTM to encoder the past and an `unrolling' LSTM to predict the future.
% ActionBanks~\cite{sener2020temporal} improves over RULSTM by carefully leveraging long-term action blocks and non-local blocks.
% AVT~\cite{girdhar2021anticipative} uses an attention-based video modelling architecture that attends to previous frames to anticipate the future.
% MeMViT~\cite{wu2022memsit}, on the other hand, processes videos online by using cache ``memory", through which the model learns to refer prior context for long-term anticipation.
% AFFT~\cite{zhong2023anticipative} improves on AVT by using multiple modalities, and using self-attention modules to fuse the features together.

\section{Contribution of Each Modality on Action Anticipation}
\begin{table*}[!t]
\centering
\small
\setlength{\tabcolsep}{6pt} % Default value: 6pt
\renewcommand{\arraystretch}{1.1}
\resizebox{0.9\linewidth}{!}{
\begin{tabular}{P{0.2cm} P{8.0cm} P{6.0cm} P{1.0cm}}
\hline
& \textbf{FasterRCNN objects} & \textbf{Actions} & \textbf{Future Action} \\ \hline
1 & `sponge, tap', `sponge, tap',`sponge, tap',`sponge, tap', `sponge, tap',`sponge, tap',`sponge, tap', `sponge, tap',`sponge, tap',`sponge, tap' & `wash plate', `wash plate', `no action', `wash plate', `wash plate', `wash plate', `wash plate', `wash plate',`insert plate', `insert plate', `wash sponge',`wash sponge',`wash sponge', `wash sponge',`wash sponge', `wash sponge' & Wash cloth \\ \hline
2 & `bin, spoon', `bin', `knife, ', `knife, ', `bin', `bin', `bag', `bag', `bag', `bin, bag', `bin, bag',`bin, bag', `bin, bag', `bin, bag', `bin, bag', `bin, bag' & `wrap bag', `wrap bag', `wrap bag', `wrap bag',`wrap bag', `wrap bag', `wrap bag', `wrap bag',`wrap bag', `wrap bag', `wrap bag', `wrap bag',`wrap bag', `wrap bag', `wrap bag' & Tie Bag \\ \hline
3 &  `cupboard', `cupboard', `cupboard', `cupboard', `cupboard',  `pan,cupboard',`pan,cupboard', `cupboard', `cupboard', `cupboard, lid',  `pan,cupboard', `pan,cupboard',`pan,cupboard', `pan,cupboard',  `pan,cupboard', `pan, ' &  `take plate', `take plate', `take plate', `take plate', `take plate', `no action', `open cupboard', `no action', `insert plate', `insert plate', `no action', `no action', `take cup',`no action', `open cupboard', `insert cup' & Put-into Cup \\ \hline
4 &  `bowl,spoon, tap, knife', `bowl,spoon, tap, knife', `bowl,spoon, tap, knife',`bowl, spoon, cup, tap, knife', `bowl, spoon, tap, knife', `bowl, spoon, tap, knife',`bowl, spoon, cup, knife, bottle', `bowl, cup, tap, knife, lid', `bowl,knife, tap',`bowl, spoon, tap, knife, lid', `bowl, spoon, tap, knife', `bowl,knife, tap',`bowl, spoon, tap, knife', `bowl, spoon, tap, knife, sponge', `knife,tap, spoon' &  `wash cup', `wash cup', `no action', `wash spoon', `wash spoon', `put spoon', `wash cup', `wash cup', `wash cup',`wash cup', `wash cup', `wash cup', `wash cup', `no action', `no action', ' turn-off tap' & Turn-off tap \\ \hline
5 &  `board:chopping, onion, knife, spatula', `board:chopping, onion, knife, spatula', `knife, onion, spatula', `knife, board:chopping, onion, spatula', `food, onion, knife, spatula',`knife, board:chopping, onion, spatula', `knife, board:chopping, onion, spatula',`knife, board:chopping,spatula', `knife, board:chopping, lid, spatula', `board:chopping,knife', `knife, board:chopping,spatula', 
% `board:chopping, knife, salad, salt, spatula', `board:chopping, knife, salt, spatula, oil', `knife, milk, salt, spatula', `knife, board:chopping,spatula', ' board:chopping, cup, salt, spatula, food'
& `put board:chopping', `no action', `no action', `put knife', `no action', `open drawer', `no action', `no action',`take spatula', `close drawer', `no action', `mix aubergine', `mix aubergine', `mix aubergine', `mix aubergine', `put aubergine' & Take salt \\ \hline
6 & `lid, glass, bottle', 'lid', 'lid, glass', 'glass, lid', `glass, lid', `glass, container', `glass, bottle, container', `bag, glass', `bag, glass', `glass, bag, bottle', `glass, bottle, oil',' glass, bag, bottle, container', `glass, bag, lid, bottle', spoon, bottle, glass, bag, lid', `onion, lid', `cup, glass, bottle, bag, lid'  & `move bin', ' move bin', `take milk', `take milk',' crush milk', `crush milk', `crush milk', `crush milk',`no action', `no action', `no action', ' insert milk',`no action', `no action', `take paper', `take paper'  & Move glass \\ \hline
\end{tabular}
}
\vspace{0.1em}
\caption{
Per frame objects and actions detected in a video clip in EPICKitchens-100 dataset. 
The objects are detected using FasterRCNN trained on EK55 dataset. 
We set a threshold of 0.15, and select only top 5 objects per frame.
Actions described here are the ground truth annotations. When actions are not detected, a `no action" label is used instead.
}
\label{tab:obj_action}
\end{table*}

In Figure \ref{fig:modalities}, we explore the contributions of various modalities to performance. 
For all the experiments, we use the objects provided by ~\cite{furnari2020rolling}, and ground truth labels for actions.

We first compare our model performance \textcolor{red}{ROFAUV} against \textcolor{bluee}{ROFA} (also noted in Table 3 of the main paper). 
We see that the additional modalities \emph{i.e} Objects and Actions significantly improve the performance.

With RGB (R) as a base modality, comparing \textcolor{yellow}{RV} with \textcolor{green1}{RUV}, we see that the objects detected using fasterRCNN model aid in the performance, however, through a small margin. 
To understand the impact of using object information as an additional modality, we examine the detected objects and the actions in Table~\ref{tab:obj_action}. 
We see that for rows 1 and 3, the object required for the action prediction is not detected by the FasterRCNN model with high probability. 
For rows 2 and 4, while the object was detected, presence of other objects make the action prediction challenging. 
On the other hand, actions (which are often defined as a verb-noun pair) give more information about the objects being interacted and the actions in the observed frames.
Therefore, while detecting objects accurately is essential and makes one part of the action ($<$verb,noun$>$), it is also vital that an active hand-object interaction be detected. 

Comparing \textcolor{green1}{RUV}, \textcolor{blue1}{FUV} and \textcolor{cyan}{AFUV} we see that audio and flow also aids in the model performance, and in combination provide the similar information to the model as the RGB data.

\clearpage
{\small
\bibliographystyle{ieee_fullname}
\bibliography{egbib}
}

\end{document}